\documentclass[11pt]{article}

\usepackage[final]{acl}
\usepackage{times}
\usepackage{latexsym}
\usepackage[T1]{fontenc}
\usepackage[utf8]{inputenc}
\usepackage{microtype}
\usepackage{inconsolata}
\usepackage{graphicx}
\usepackage[belowskip=-0.3em,labelfont=bf,labelsep=endash]{caption}
\usepackage[table]{xcolor}
\definecolor{lightgray}{gray}{0.9}
\usepackage{booktabs}
\usepackage{amsmath}
\usepackage{amssymb}
\usepackage{mathtools}
\usepackage{amsthm}
\usepackage{xspace}
\usepackage{newtxtext}
\usepackage{scalerel}

\newcommand{\faGitLogo}{\scalebox{1.2}{\scalerel*{\includegraphics{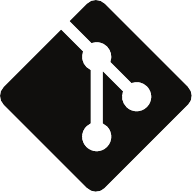}}{X}}}

\newcommand{\lagom}{L\textsuperscript{\hspace{-3pt}A}G\textsubscript{O}M$\boldsymbol{\cdot}$NLP}

\newcommand{\bo}{\texttt{bo}\xspace}
\newcommand{\dz}{\texttt{dz}\xspace}
\newcommand{\nep}{\texttt{ne}\xspace}
\newcommand{\my}{\texttt{my}\xspace}
\newcommand{\zho}{\texttt{zh}\xspace}
\newcommand{\khm}{\texttt{km}\xspace}

\newcommand{\tibt}{\texttt{Tibt}\xspace}
\newcommand{\mymr}{\texttt{Mymr}\xspace}
\newcommand{\deva}{\texttt{Deva}\xspace}

\newcommand{\scnum}[1]{\scalebox{0.8}{#1}}

\newcommand{\pixel}{\textsc{pixel-mono}\xspace}
\newcommand{\pixelm}{\textsc{pixel-m\scnum{4}}\xspace}

\usepackage[table]{xcolor}
\definecolor{pixelblue}{RGB}{31,119,180}

\title{Seeing the Unseen: Visual Similarity for Pixel Language Model Adaptation}

\author{Ran Zhang \quad Miryam de Lhoneux \quad Wessel Poelman\\
    \lagom, Department of Computer Science, KU Leuven\\ 
    \texttt{\{firstname.lastname\}@kuleuven.be}
}

\begin{document}
\maketitle

\begin{abstract}
    Pixel-based language models (LMs) replace traditional tokenizers by processing rendered images of text, making cross-lingual transfer heavily dependent on the visual and structural properties of writing systems. 
    However, the dynamics of adapting these models to low-resource languages with complex morphology and written in unique scripts are not yet explored.
    Using Tibetan as a case study, we analyze how continued pre-training of pixel-based LMs is influenced by data scale, initial script exposure, and cross-lingual transfer from languages written in other Brahmic scripts.
    We introduce four rendering-level metrics to quantify visual script similarity.
    We evaluate downstream performance across three tasks.
    Our results show that higher orthographic proximity enhances semantic transfer, even under severe data constraints. Additionally, we find a performance asymmetry based on the pre-training starting point: 
    while multilingual pre-training (\pixelm) has stronger initial performance, its capacity for subsequent adaptation seems to be constrained, whereas adapting a monolingual model (\textsc{pixel}) with mixed scripts yields more gains on sentence-level tasks. 
    Our metrics and case study offer empirical observations that could help inform data selection and script adaptation choices when working with pixel-based models in similar low-resource settings.
\end{abstract}

\vspace{-0.75em}
\begin{center}
    \faGitLogo\ \href{https://github.com/RAN-rz/seeing_the_unseen}{\texttt{RAN-rz/seeing\_the\_unseen}}
\end{center}

\section{Introduction}\label{Introduction}
Pixel-based language models (LMs) are a promising direction for multilingual language modeling.
Instead of processing tokens, these models process rendered images of text.
Subword-based models perform generally well on high-resource languages but often struggle with low-resource languages \cite{mielke2021words}.
Languages written in Non-Latin scripts are often not only low-resource, but are also hindered by requiring more Unicode code-points to represent \cite{petrov2023language,vandergoot2024where}. 
``Tokenizer-free'' models do language modeling over units that are potentially fairer across languages, such as characters or bytes \cite{clark2022canine,xue2022byt5}.
A drawback of such models is their computational efficiency since they need to process long input sequences.
Pixel-based languages models \cite{salesky2021robust,rust2023language,kesen2025multilingual} circumvent such limitations by rendering text as sequences of image patches.
This is promising for multilingual models since cross-lingual transfer can potentially happen purely based on \emph{visual similarity} of writing systems \cite{rahman2023token}.
With subword-based models, transfer generally happens through token overlap across similar languages.
Across scripts, this is nearly impossible, say for named entities or loan words.
With pixel-based models, this restriction is potentially removed if we use visually similar scripts.
However, there is no clear definition or measure of ``visual similarity'' in previous work, apart from broad categories like Similar and Not-Similar or if writing systems belong to the same family (Brahmic for example).
A more fine-grained definition would help in choosing transfer languages, which is especially important for low-resource languages since pre-training a model from scratch is often not feasible due to data limitations.

Existing work on pixel language models is generally done on high-resource languages:
the original \textsc{pixel} model \cite{rust2023language}, hereafter \pixel, is only trained on English, the subsequent \pixelm model \cite{kesen2025multilingual} uses four high-resource languages (English, Chinese, Ukrainian, and Hindi), \textsc{pixar} \cite{tai2024pixar} is a monolingual English model, and \textsc{mixar} \cite{hu2026mixar} uses German, English, Spanish, Italian, French, Korean, Chinese, and Japanese.
It is hard to disentangle the effect of \emph{visual similarity} since the languages in \pixelm are chosen with visual \emph{diversity} in mind, and in \textsc{mixar} there is a lot of existing script overlap.

We address two research gaps: (1) a more fine-grained definition of \emph{visual similarity}, and (2) a detailed analysis of low-resource adaptation for pixel language models. 
We focus on Tibetan as a case study.
Tibetan not only suffers from a lack of available resources,\footnote{Tibetan is classified as the second-lowest resource tier ``Scraping By`` according to \citet{joshi2020state}.} but also possesses unique orthographic and morphological features. 
Tibetan is generally written in the Tibetan script (\tibt), which is an abugida writing system belonging to the Brahmic family.
Syllable clusters are the basic unit of writing.
Consonant serve as the root, to which superscript and subscript and vowels may be attached; syllable clusters are separated by \emph{tsheg} (syllable delimiters) rather than spaces \cite{wang2023effects}. 
This syllabic structure, characterized by multiple consonant combinations, results in Tibetan glyphs having high visual density and complex structures, making it significantly different in style from related scripts from the Brahmic family.
At the morphological level, Tibetan features a rich agglutinative morphology with complex affixation \cite{lai2018tibetanchinese,li2022characterbased}, posing additional potential challenges for modeling. 
Collectively, these features make Tibetan a good fit for our case study.
Our contributions are:

\begin{itemize}
    \item \textbf{Quantifying visual similarity:} we propose four metrics for ``visual similarity'' based on the rendered patches the model processes, useful both for analyzing pixel model behavior across scripts and for guiding language selection in cross-lingual transfer.
    
    \item \textbf{Exploring the feasibility of continued pre-training (CPT) for pixel-based models:} there is no existing work on adapting pre-trained Pixel models to low-resource languages written in new scripts. We explore this using Tibetan as a case study.
    
    \item \textbf{Quantifying the data requirements in adapting Pixel language models:} we run ablation experiments to see how ``low-resource'' we can go with CPT for adapting a pixel-based model to an unseen script.
\end{itemize}

\section{Related Work}\label{sec:related-work}
\paragraph{Tokenization.} Currently, mainstream LMs tend to rely on subword tokenizers, such as BPE \cite{sennrich2016neural} or UnigramLM \cite{kudo2018subword}. They have been proven to work well on high-resource languages, but bring a structural bottleneck for low-resource languages due to fixed vocabulary sizes and training data availability \cite{mielke2021words}. Low-resource languages are often disadvantaged, making out-of-vocabulary (OOV) issues more severe \cite[\emph{e.g.,}][]{abbott-martinus-2019-benchmarking,bostrom2020byte,mielke2021words}. This is often worse for languages written in non-Latin scripts \cite{petrov2023language,vandergoot2024where}. Pixel-based models are therefore attractive for such languages, as they sidestep tokenization. 

\paragraph{Tokenization-free.} To overcome the limitation of a finite vocabulary, tokenizer-free methods have been proposed \cite{mielke2021words}. These include LMs trained directly on characters (CANINE, \citealt{clark2022canine}) or bytes (ByT5, \citealt{xue2022byt5}), with the promise of providing a unified and fair input space across languages. Another approach is to represent text as images \cite{salesky2021robust}, removing the finite vocabulary of a tokenizer and instead relying solely on sequences of image patches. \pixel \cite{rust2023language} is an English-only encoder model that uses a Vision Transformer \cite{dosovitskiy2020image}. \pixelm \cite{kesen2025multilingual} extends this idea to a multilingual setting, using four languages. \textsc{pixar} \cite{tai2024pixar} is an English, decoder-only variation. \textsc{mixar} \cite{hu2026mixar} builds on top of it and extends its language coverage. Collectively, these pixel-based models demonstrate promising cross-script transfer across diverse languages and scripts, and the unique advantages of multilingual modeling with pixels.

\begin{figure*}[ht]
    \centering
    \includegraphics[width=\linewidth]{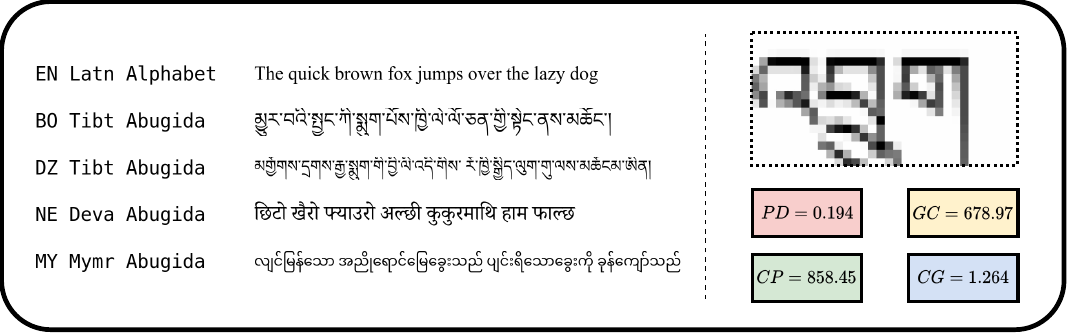}
    \caption{We compare four settings of \emph{script similarity}: in the example above, all writing systems are abugidas belonging to the Brahmic family (except English). The goal of our metrics is to capture similarity on a finer-grained level and with units the model actually processes. The visual example on the right is a single patch, in actual use our metrics are calculated over the entire input blocks.}
    \label{fig:diagram}
\end{figure*}

\paragraph{Transfer and Visual Similarity.} Cross-lingual transfer has been researched for subword-based models as a way to address the performance gap between high- and low-resource languages \cite[\emph{e.g.,}][]{lauscher2020zeroa,philippy2023common,muller2023languages}. Linguistic relatedness is one of the contributing factors \cite{lin2019choosing, ogueji2021small}, which is even more critical in low-resource settings than increasing the scale of data \citep{devries2021adapting,ogunremi2023mini}. A principled selection of multilingual data can boost performance for low-resource languages, but once the number of languages and data reach a certain size, both low- and high-resource languages may face the curse of multilinguality \cite{chang2024when}.

For pixel-based models, transfer differs as inputs are not tokens but visual patches. Thus, cross-lingual transfer is more likely to be influenced by visual similarities \cite{rahman2023token}. Both \pixel \cite{rust2023language} and \pixelm \cite{kesen2025multilingual} showed solid cross-script transfer to unseen languages. However, this transfer mechanism has not been explored; existing research often equates \emph{visual similarity} to script similarity, or simply classifies such similarity to binary features. \citet{roman2026contrastivetoselfsupervised} point out that cross-script similarity between writing systems should not be reduced to discrete categories. Differences within the same script family are continuous and gradient.
Addressing this is valuable both as an analytical tool for rendering and pixel-based models and as a practical guide for selecting languages in cross-lingual transfer.

\paragraph{Tibetan.} We select Tibetan as our central language for several reasons. First, Tibetan is a typical case for the aforementioned situations, featuring complex morphology and limited available data resources (see \S\ref{Introduction}). While wide-coverage, tokenizer-based models such as Glot500 \cite{imanigooghari2023glot500}, and NLLB \cite{costa-jussa2024scaling} include Tibetan, it has generally received little attention in NLP research. Some specialized approaches include: CINO \cite{yang2022cino}, an encoder pre-trained with Tibetan and seven other languages; TiBERT \cite{liu2022tibert}, the first monolingual Tibetan LM, and some task specific studies such as Tibetan named entity recognition \cite{barnett2021named} and Tibetan-Chinese machine translation \cite{lai2018tibetanchinese}. Second, the Tibetan script (\tibt), belongs to the Brahmic script family, sharing typological roots with Hindi, used for \pixelm \cite{kesen2025multilingual} pre-training. All in all, Tibetan serves both as a language relatively overlooked by mainstream NLP systems and as a suitable test case for examining the \emph{visual similarity} mechanisms that remain ``unseen'' in pixel-based models.

\section{Methodology}\label{sec:methodology}
We first introduce our \emph{visual similarity} metrics, discuss how it behaves across languages and scripts, and finally outline our experimental setup.

\subsection{Visual Similarity Metrics}\label{sec:Visual Similarity Metrics}
Language similarity can be defined along many axes.
For pixel-based models, we cannot readily use existing token or lexical measures.
Our metric needs to be applicable to \emph{patches} (long rendered images of text) and relevant to studying cross-lingual transfer.
While we could use broad categories, like other abugidas or the Brahmic family, there has been a push for more fine-grained analyses of languages and model behavior beyond coarse typological groupings \cite{poelman2025confounding}.

\paragraph{Definitions.} We define four metrics (see Figure~\ref{fig:diagram} for an overview) that characterize the orthographic similarity between languages on the patch level.
We calculate these based on pre-rendered data for each language. Let $N$ be the total number of blocks in the data set, $T$ be the fixed number of patches per block, $H = W = 16$ the height and width of each patch in pixels, and $x_{i,j,k} \in [0, 255]$ the value of the $k$-th pixel in the $j$-th patch of block $i$, where $255$ represents a white background. Let $c_i$ be the number of Unicode code-points in block $i$, and $g_i$ the number of grapheme clusters, human-perceived characters under Unicode text encoding in block $i$.

\textit{Pixel density} measures the proportion of non-background pixels across all blocks, providing a  proxy for ink coverage and glyph complexity:
\begin{equation}
    PD = 
    \frac{\displaystyle\sum_{i=1}^{N} \sum_{j=1}^{T} \sum_{k=1}^{H \times W} 
    \mathbb{1}[x_{i,j,k} < \tau]}{N \times T \times H \times W}
    \label{eq:pixel_density}
\end{equation}

\noindent where $\tau = 250$ is the background threshold, retaining a 
small margin to account for anti-aliasing artifacts.

\textit{Average grapheme clusters per block} reflects the number of human-perceived characters the renderer encodes per input block:
\begin{equation}
    GC = \frac{1}{N} \sum_{i=1}^{N} g_i
    \label{eq:avg_gc}
\end{equation}

\textit{Average code-points per block} captures the raw Unicode encoding length of each input block:
\begin{equation}
    CP = \frac{1}{N} \sum_{i=1}^{N} c_i
    \label{eq:avg_cp}
\end{equation}

\textit{The code-points-to-grapheme ratio} reflects the degree of glyph compositionality: scripts with frequent stacking or conjunct forms require multiple code-points per perceived character, yielding a higher ratio:
\begin{equation}
    CG = \frac{CP}{GC}
    \label{eq:cp_gc_ratio}
\end{equation}

\paragraph{Metric behavior.} We apply our metric to 41 languages.
We use the Parallel Bible Corpus \cite[PBC;][]{mayer-cysouw-2014-creating} written in 23 different scripts.
The corpus is aligned by verses and rendered into patches (see \S\ref{sec:Rendering Details} for details).
Next, we create vectors per language that consist of the four metrics. 
When looking at the Euclidean distances between vectors, we see an interesting pattern: both the \emph{closest} and \emph{most distant} language to Tibetan come from the Brahmic family: Khmer (\texttt{Khmr}) and Burmese (\texttt{Mymr}), respectively.
Table~\ref{tab:visual_similarity_41langs} in \S\ref{apx:full-results} contains the full results for all metrics.
A PCA projection of the resulting language vectors \S\ref{apx:41 lang}.
We also show that clustering of these vectors can recover coherent script groups.

The metrics measure properties of the rendered patches that the model also processes.
We next investigate whether this is helpful in explaining cross-lingual transfer in pixel-based models.

\subsection{Language Selection}\label{sec:language selection}
We select a language sample for our continued pre-training experiments. We first select three other low-resource languages\footnote{All are in the ``scraping by'' category \cite{joshi2020state}.} that share broad categories with \textbf{Tibetan} (\bo): \textbf{Dzongkha} (\dz) is a dialect of Tibetan, also commonly written in the Tibetan script (\tibt), therefore being an abugida, and belonging to the Brahmic family. This represents a slight variation of Tibetan, written in the same script, but being mutually unintelligible \cite{book}. \textbf{Nepali} (\nep) is normally written in Devanagari (\deva), which is also an abugida, and belongs to the Brahmic family. \textbf{Burmese} (\my) is usually written in another Brahmic abugida: Myanmar (\mymr). Since Dzongkha is not included in the PBC, it could not be incorporated into the validation of the 41 languages mentioned above. Its inclusion reflects the fact that, like Tibetan, it is one of the few languages still written in Tibetan script. Figure~\ref{fig:diagram} shows the same sentence in the different languages to get an idea of the similarities and differences between the scripts.

Among the 41 languages from the PBC, \textbf{Khmer} (\texttt{Khmr}) is the language closest to Tibetan, Burmese (\texttt{Mymr}) is the farthest, and \textbf{Chinese} (\texttt{Hans}) the second farthest. We supplement our main experiments with a smaller follow-up study, using the two languages at opposite ends of the distance distribution: Khmer and Chinese, since Burmese is already included in our main case study language sample.

\begin{table}[h]
    \centering
    \footnotesize
    \setlength{\tabcolsep}{4pt}
    \resizebox{\columnwidth}{!}{

\begin{tabular}{cc|ccccc}
\toprule
\textbf{Lang} & \textbf{Script} & $PD$ & $GC$ & $CP$ & $CG$ & \textbf{Dist.} \\
\midrule
\texttt{bo} & \texttt{Tibt} & 0.19 & 678.97 & 858.45 & 1.26 & -- \\
\texttt{dz} & \texttt{Tibt} & 0.19 & 636.83 & 815.11 & 1.28 & 0.52 \\
\texttt{ne} & \texttt{Deva} & 0.16 & 502.89 & 782.18 & 1.56 & 2.32 \\
\texttt{my} & \texttt{Mymr} & 0.14 & 378.02 & 562.08 & 1.49 & 3.97 \\
\texttt{zh} & \texttt{Hans} & 0.25 & 608.31 & 608.31 & 1.00 & 2.70 \\
\texttt{km} & \texttt{Khmr} & 0.21 & 541.49 & 874.24 & 1.62 & 2.02 \\
\bottomrule
\end{tabular}
}
    \caption{Visual similarity metrics on the patch level. For \bo, \dz, \nep and \my, distances are based on pre-training data; for \zho and \khm, distances are based on Table~\ref{apx:41 lang}. The rankings are consistent between the two.}
    \label{tab:visual_similarity_six}
\end{table}
Table~\ref{tab:visual_similarity_six} shows how our metrics capture the degree of similarity between the languages written in the different scripts.
Dzongkha is indeed at the high end (sharing the \tibt script), Nepali in the middle range, and Burmese at the lowest end, which both confirms our intuition of these scripts and aligns with their relative positions in the 41-language ranking Table~\ref{tab:visual_similarity_41langs} (see \S\ref{sec:auxiliary-language} for a detailed analysis).

Quantitatively, $PD$ reflects coverage and visual complexity of glyphs. It is nearly identical for Tibetan (0.194) and Dzongkha (0.191), while  Nepali (0.160) and Burmese (0.138) are notably lower. 
$GC$ (Eq.~\ref{eq:avg_gc}) captures the number of human-perceived characters encoded by the renderer for each block. It is substantially higher for Tibetan (678.97) and Dzongkha (636.83) than for Nepali (502.89) and Burmese (378.02), suggesting that the former two are also more similar in block level rendering granularity and character density. 
$CP$ reflects the length of the Unicode encoding and follows the same pattern: Tibetan (858.45) and Dzongkha (815.11) are considerably higher than Nepali (782.18) and Burmese (562.08), further confirming that Dzongkha is structurally closed to Tibetan. 
Lastly, $CG$ measures the complexity of glyph combinations as the average number of Unicode code-points corresponding to per human-perceived character. 
It is higher for Nepali (1.555) and Burmese (1.487) than for Tibetan (1.264) and Dzongkha (1.280), implying that for Nepali and Burmese, more code points are required to form a perceptible character. 
This structural difference at the encoding level further proves the greater orthographic distance between these two languages and Tibetan. 
Across all four metrics, orthographic proximity between the three languages also written in Brahmic scripts and Tibetan decreased in the following order: \mbox{\dz $>$ \nep $>$ \my}, consistent with the qualitative analysis above. Khmer (\texttt{Khmr}) and Chinese (\texttt{Hans}), the two languages selected for the follow-up experiments, exhibit similarity and divergence. Khmer's $PD$ is 0.207 and $CG$ is 1.615 both fall in the same high range as Tibetan ($PD$ = 0.202, $CG$ = 1.300), consistent with it being the closest language to Tibetan in the 41-language ranking. Chinese, by contrast, has a notably higher $PD$ (0.247) but a markedly lower $CG$ (1.000), reflecting Chinese characters as self-contained logographic units without the multi-codepoint stacking structure of Tibetan. It suggests that its distance from Tibetan is not uniform across dimensions but rather reflects divergence of varying degree and direction across metrics, further indicating that the four metrics work in combination rather than in isolation, see \S\ref{apx:41 lang} for a more detailed analysis.

\paragraph{Metric consistency.} The values for Tibetan, Dzongkha, Nepali and Burmese in Table~\ref{tab:visual_similarity_six} are computed on the same rendered corpora used for the continued pre-training (see \S\ref{sec:data} for data sources). In contrast, the values for Chinese and Khmer in Table~\ref{tab:visual_similarity_six} are computed on the PBC \cite{mayer-cysouw-2014-creating}. Table~\ref{tab:visual_similarity_41langs} reports the corresponding PBC-based values for both \bo, \nep and \my as well. Since Dzongkha is not included in the PBC, it cannot be compared across corpora. For Nepali and Burmese, however, despite the substantial domain differences between web-crawled and religious text, their similarity ranking relative to Tibetan remains consistent across both corpora. Burmese is consistently ranked farther from Tibetan than Nepali. Absolute values fluctuate to some degree, most noticeably for Burmese, suggesting that the absolute readings of these metrics may carry some sensitivity to text domain. 

\subsection{Data}\label{sec:data}
\paragraph{Continued pre-training.} We want to see how well the pixel-based models can be adapted to Tibetan and how the potential transfer languages can help.
There is a shortage of high-quality open-source Tibetan text resources; we start with a total of 714k sentences from OPUS \cite{tiedemann2012parallel} and NLLB \cite{costa-jussa2024scaling}. Dzongkha resources are even scarcer; we collected approximately 97k sentences from FineWeb2 \cite {penedo2025fineweb2a} and OPUS. The Nepali and Burmese data are both sourced from OPUS and NLLB, with approximately 330k sentences randomly sampled from each corpus. The 41 languages used to validate the visual similarity metrics are drawn from the Parallel Bible Corpus \cite[PBC;][]{mayer-cysouw-2014-creating}. For the follow-up continued pre-training experiment, Chinese and Khmer data are sourced from FineWeb2, with approximately 420k and 175k sentences respectively.  

All corpora are normalized before rendering: UTF-8 re-encoding, removing surrogate and private-use Unicode characters, filtering non-readable characters following \pixelm \cite{kesen2025multilingual}.\footnote{\pixelm \href{https://github.com/ilkerkesen/pixel-m4/blob/main/scripts/data/unrenderable_chars.json}{unrenderable character list}.} Additionally, language-specific filtering is applied: sentences with low langid scores ($< 0.80$) are removed \cite{lui-baldwin-2012-langid}, emoji and Latin characters are filtered, and URLs are removed.

All corpora were rendered using \mbox{PangoCairoTextRenderer},\footnote{Using the \href{https://fonts.google.com/noto}{Noto Sans} font.} with each patch measuring 16x16 pixels. 
All data is rendered continuously on the sentence level, instead of the bi-gram rendering strategy \cite{lotz2023text} adopted by \pixelm \cite{kesen2025multilingual}, see \S\ref{sec:Rendering Details} for a detailed explanation. To standardize the corpus sizes across different languages and facilitate controlled comparison, the full Tibetan corpus (35k blocks) is used after rendering, and the other three languages are scaled to approximately 24k blocks to match the rendering results of Dzongkha, which has the smallest dataset. In all subsequent experiments, data scale is measured in rendering blocks\footnote{One block contains approximately 22 sentences.} rather than sentence count or token count, to align with Pixel-based models' input units as rendered patches.
\paragraph{Downstream tasks.}
To evaluate the effectiveness of CPT, we evaluate on three Tibetan downstream tasks, covering both semantic and syntactic levels: Text Classification (12 categories) sourced from the Tibetan News Classification \cite[TNCC;][]{qun2017end}, NER (5 entity types) sourced from the Tibetan and Mongolian Newspaper NER dataset \cite{barnett2021named}, and POS tagging (17 UPOS tags) sourced from the Tibetan dependency treebank \cite{quecairang2013treebank}. 

To our knowledge, this constitutes the first evaluation of pixel-based LMs on Tibetan, combining rendering with native Tibetan data across these three tasks. All downstream data are pre-rendered using the same strategy as the CPT corpora, but with different sequence lengths, following the configurations of \pixel \cite{rust2023language} and \pixelm \cite{kesen2025multilingual}. Macro-averaged F1 is used as the primary evaluation metrics across all tasks; for POS tagging, we report adjusted macro-F1 excluding four extremely low-frequency tags.\footnote{Namely: \texttt{CCONJ}, \texttt{INTJ}, \texttt{SCONJ}, \texttt{SYM}.} Full dataset statistics and preprocessing are provided in \S\ref{sec:Rendering Details}. 

\subsection{Experiments}\label{sec:experiments}
We conducted CPT using \pixel \cite{rust2023language} and \pixelm \cite{kesen2025multilingual} as two independent starting points to examine the impact of monolingual and multilingual pre-training on subsequent Tibetan language adaptation. In the data scale experiments, we set up six CPT scale conditions based on the Tibetan corpus: no CPT, +1K, +4K, +8K, +24K and +35K blocks, to explore data scale sensitivity under very-low-resource conditions. In the auxiliary language experiments, all languages use approximately 24k blocks for CPT to ensure comparability of scale. 

We follow the pre-training and fine-tuning recipe of \pixel and \pixelm; detailed hyperparameter configurations are provided in \S\ref{apx:training-details}.

We define several types of baselines: pixel-based, character-based, and subword-based. The latter two comparison models are included as reference points rather than strictly comparable baselines. Both these models were pre-trained on substantially more languages and data than the models considered here.

\paragraph{(1) Pixel-based.} We use \pixel \cite{rust2023language} and \pixelm \cite{kesen2025multilingual} without CPT as pixel-based baselines. Finetuning for downstream tasks is conducted directly on the original pre-training weights. \pixel was pre-trained exclusively on English, without any exposure to any language similar to Tibetan in terms of script. \pixelm, was pre-trained with four languages: English, Chinese, Ukrainian, and Hindi. Hindi, written in Devanagari, also belongs to the Brahmic writing system like Tibetan and thus shares a certain degree of orthographic proximity. Therefore, \pixel offers a cleaner zero-proximity baseline, allowing the cross-lingual transfer experiments with auxiliary languages to be presented more clearly. \pixelm, on the other hand, reflects adaptation behavior within a multilingual pre-training context. A comparison of the two can also reveal the potential influence of script exposure during pre-training phase on subsequent CPT performance.
Training pixel-based models on too many scripts could be harmful (curse of multilinguality) or beneficial for transfer.

\paragraph{(2) Character-based.} CANINE-S \cite{clark2022canine} is taken as a character-level point of comparison. The model was pre-trained on a multilingual Wikipedia corpus with 104 languages. For all the downstream tasks, we employ the same fine-tuning hyperparameter settings as the PIXEL models, see details in \S\ref{apx:fintuning-details}. 
\paragraph{(3) Subword-based.} Glot500-base \cite{imanigooghari2023glot500} is taken as the multilingual subword-based point of comparison. As one of the most linguistically diverse LMs, Glot500-base covers over 500 languages for pre-training, Tibetan included, serving as a strong benchmark for subword-based methods on low-resource languages. The fine-tuning settings follow those of the PIXEL models, see details in \S\ref{apx:fintuning-details}.

\section{Results and Analysis}\label{sec:results}
\subsection{Continued Pre-training Results}
Table~\ref{tab:pixel-m4-results} shows the results from our CPT experiments.
In terms of base performance without CPT, \pixelm noticeably outperforms \pixel across all the three tasks (NER: 69.08 vs 59.49, POS: 70.57 vs 70.46, Text Classification: 44.34 vs 41.73), indicating that multilingual pre-training of \pixelm has benefited Tibetan-related tasks. This advantage may be partially attributed to Hindi used in pre-training. The Devanagari script of Hindi also belongs to the Brahmic writing system as Tibetan, and shares a certain degree of orthographic proximity, which appears to aid cross-lingual transfer.

\begin{table}[t]
\centering
\small
\setlength{\tabcolsep}{4pt}
\resizebox{\linewidth}{!}{\begin{tabular}{lccc}
\toprule
\textbf{Condition} & \textbf{NER} & \textbf{POS} & \textbf{TextCL} \\
\midrule
\pixel (no CPT)              & 59.49{\scriptsize $\pm$0.11}          & 70.46{\scriptsize $\pm$0.23}          & 41.73{\scriptsize $\pm$0.20} \\
\pixelm (no CPT)           & 69.08{\scriptsize $\pm$0.02}          & 70.57{\scriptsize $\pm$0.85}          & 44.34{\scriptsize $\pm$0.53} \\
\addlinespace[3pt]
\pixelm + \bo 35k           & {70.19}{\scriptsize $\pm$1.28} & 71.44{\scriptsize $\pm$0.11}          & 47.72{\scriptsize $\pm$0.81} \\
\pixelm + \bo 35k + \dz 24k  & 69.68{\scriptsize $\pm$0.66}          & \textbf{71.76}{\scriptsize $\pm$0.20} & 47.48{\scriptsize $\pm$0.14} \\
\pixelm + \bo 35k + \my 24k  & 68.26{\scriptsize $\pm$1.80}          & 71.15{\scriptsize $\pm$0.54}          & 46.34{\scriptsize $\pm$2.40} \\
\pixelm + \bo 35k + \nep 24k  & 69.38{\scriptsize $\pm$0.57}          & 71.26{\scriptsize $\pm$0.48}          & 47.75{\scriptsize $\pm$0.13} \\
\pixelm + allmixed 107k
                             & \textbf{70.55}{\scriptsize $\pm$0.02} & 71.72{\scriptsize $\pm$0.12} & 48.61{\scriptsize $\pm$0.47} \\
\addlinespace[2pt]
\pixel + allmixed 107k
                             & 68.26{\scriptsize $\pm$0.41}          & 71.41{\scriptsize $\pm$0.09}          & \textbf{54.57}{\scriptsize $\pm$0.39} \\
\midrule[\heavyrulewidth]
\textcolor{gray}{Glot500}   & \textcolor{gray}{82.59} & \textcolor{gray}{72.47} & \textcolor{gray}{62.14} \\
\textcolor{gray}{CANINE-S}  & \textcolor{gray}{78.72} & \textcolor{gray}{77.10} & \textcolor{gray}{52.13} \\
\bottomrule
\end{tabular}

}
\caption{Results for CPT. All metrics report macro-averaged F1. Results are over three random seeds. Bold indicates best performance among PIXEL-based models. allmixed comprises \bo + \dz + \my + \nep (${\sim}$107k blocks total). The subword- and character-based models are included for reference, but are not directly comparable.}
\label{tab:pixel-m4-results}
\end{table}

Under the Tibetan-only (\bo) CPT condition, semantic tasks again show noticeable gains (NER: +1.11; Text Classification: +3.38), while POS tagging remains largely unchanged (we discuss this more in \S\ref{sec:sem-analysis}). Compared to \pixel, the absolute gains across tasks for \pixelm are generally smaller; the previously discussed inclusion of Hindi may be one of the causes of this. 

The three pairwise mixing conditions, \bo (35k) + \dz (24k), \bo (35k) + \my (24K) and \bo (35k) + \nep (24K) use identical data ratio yet yield limited performance differences across all three tasks, with neither consistently outperforming the \bo only condition. This suggests that at the current data scale, the impact of orthographic proximity of auxiliary languages may be masked by other factors, or may not yet have reached the scale threshold required to yield consistent benefits. This contrasts with finding in \S\ref{sec:auxiliary-language}, where the CPT in auxiliary languages brings positive cross-lingual transfer for \pixel \cite{rust2023language}. The divergence likely stems from differences in pre-training coverage: for \pixel, the 24k blocks of auxiliary language data constitute the primary source of non-English (non-Latin) text exposure for the model. However, the \pixelm pre-training data scale is aligned with that of mBERT \cite{devlin2019bert}, with each of the four languages receiving a substantial volume of pre-training data.
\citet{kesen2025multilingual} use an estimated 50 million sentences per language, which is approximately 2 million blocks. 
Our 24k blocks of auxiliary language data introduced during CPT account for a relatively limited proportion of the overall representation space.

The allmixed condition (\bo + \dz + \my + \nep, 107 blocks) achieves the highest scores among all \pixelm configurations on Text Classification (48.61) and NER (70.55), but underperforms the \bo + \dz condition on POS (71.72 vs 71.76). Comparing the corresponding \pixel + allmixed configuration reveals an asymmetry: \pixel + allmixed achieves 54.57 on Text Classification, substantially higher than \pixelm + allmixed (48.61), whereas on NER (68.26) it falls below \pixelm (70.55), with POS scores remaining close (71.41 vs 71.72). This inconsistency implies that the interaction between multilingual pre-training and large-scale mixed CPT data is moderated by task type, see details in \S\ref{sec:sem-analysis}.

\paragraph{Comparison Models.}
We compare the best-performing PIXEL configuration with two larger-scale models. These comparisons are not directly fair due to much bigger model sizes and amount of pre-training data. We include these to have a point of reference for where Pixel-based models stand. Glot500 (subword-based) outperforms all PIXEL model configurations across all the three tasks (NER: 82.59, POS: 72.47, Text Classification: 62.14), demonstrating that large-scale multilingual pre-training with direct target-language coverage can yield impressive performance even for low-resource languages. Its pre-training corpus includes Tibetan, and its scale, with 511 languages, 600GB data, 1.225 billion sentences, far exceeds that of all models in this study.

CANINE-S (character-based) outperforms all PIXEL configurations on NER (78.72) and POS (77.10). However, on Text Classification (52.13), CANINE-S performs comparably to \pixel \cite{rust2023language} with 35k Tibetan CPT (52.75) and 24k Dzongkha CPT (53.41). Given that CANINE-S is pre-trained on 104 languages, the ability of pixel-based models to match or exceed it on sentence-level semantic tasks suggests that leveraging orthographic proximity may be an effective complementary approach for semantic transfer under low-resource settings. 

\subsection{Visual Similarity}\label{sec:auxiliary-language}
To test the influence of visual similarity on cross-lingual transfer, we use a fixed amount of 24k rendered blocks for each of our three transfer languages and scripts.
As discussed in \S\ref{sec:Visual Similarity Metrics}, the Tibetan script is characterized primarily by horizontal and vertical stacking structures \cite{lai2018tibetanchinese,li2022characterbased}, where superscripts and subscripts
form locally dense, block-wise clusters that visually appear square and compact. Dzongkha shares the \tibt script with Tibetan, and closely mirrors Tibetan in overall typographic style, making it the most orthographically proximate of the  three languages. Nepali (\nep) uses the Devanagari script with relatively consistent character widths and a similar structure of stacked consonants and thus is visually similar to Tibetan in terms of character shape and overall density. However, the śirorekha (continuous horizontal headline) distinguishes it from Tibetan's block-wise layout, positioning Nepali in the middle range of orthographic proximity. Burmese, written in the Myanmar script, is characterized by round curves, with few acute angles of square structures, making it the most visually distant from Tibetan among the three. Beyond the three languages, we additionally incorporate two languages positioned at the extremes of the orthographic continuum, based on the quantitative validation over 41 languages as mentioned in \S\ref{sec:Visual Similarity Metrics}: Khmer, the closest of all 40 candidate languages to Tibetan, and Chinese, the second most distant, since the most distant language, Burmese, is already included in the primary language set. 

\begin{figure}[t]
    \centering
    \includegraphics[width=\columnwidth]{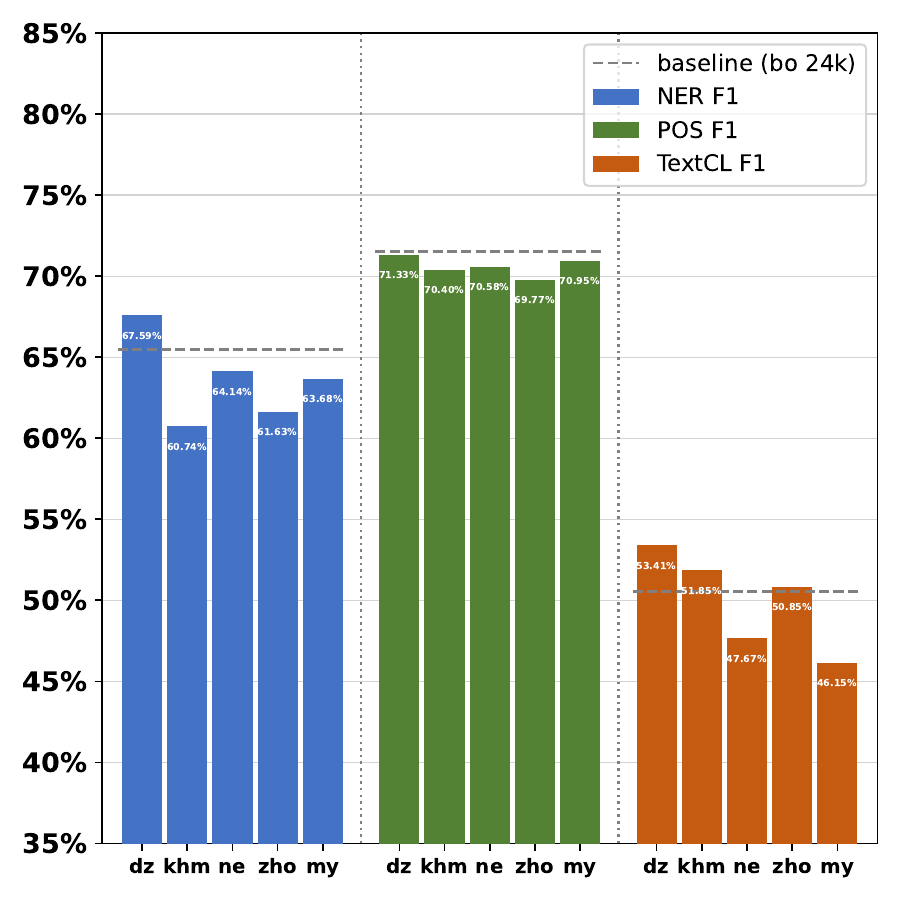}
    \caption{Effect of auxiliary language orthographic proximity on downstream task performance.
    Each bar represents a \pixel model with CPT on 24k blocks of the
    indicated auxiliary language. The dashed line indicates the \bo~24k baseline (Tibetan self-CPT).
    Languages are ordered left to right by decreasing orthographic proximity to Tibetan: Dzongkha (\dz) $>$ Khmer (\khm) $>$ Nepali (\nep) $>$ Chinese (\zho) $>$ Burmese (\my).}
    \label{fig:related-lang-transfer}
\end{figure}

For text classification, performance follows a clear gradient mirroring the orthographic proximity ranking: \dz (53.41) $>$ \nep (47.67) $>$ \my (46.15), suggesting that \emph{visual similarity} to Tibetan is a contributing factor in semantic transfer. Notably, \dz even outperforms Tibetan self-CPT baseline (50.56) on Text Classification, likely due to the higher quality of the Dzongkha corpus, which is predominantly sourced from official government and financial reports, resulting in greater lexical diversity. NER shows a similar trend, consistent with \dz (67.59) outperforming both \nep (64.14) and \my (63.68), consistent with its highest orthographic proximity to Tibetan. POS tagging shows minimal variation across all three conditions, as further discussed in \S\ref{sec:sem-analysis}. 

The follow-up languages, Khmer and Chinese, exhibit different patterns. Chinese's performance broadly aligns with its position as the second most distant language on the orthographic proximity spectrum: its scores across the three tasks (TextCL: 50.85, NER: 61.63, POS: 69.77) are generally no higher than those of Dzongkha. Khmer, by contrast, presents a more pronounced counterexample. Despite being the closest language to Tibetan aside from Dzongkha, it has the lowest NER score among the five languages (60.74), even below the most distant, Burmese (63.68). This counterexample suggests that the effect of orthographic proximity on downstream performance may be task-dependent, as discussed in \S\ref{sec:Data Size Abl}.

\subsection{Data Size Ablation}\label{sec:Data Size Abl}
\begin{figure}[t]
  \centering
  \includegraphics[width=\linewidth]{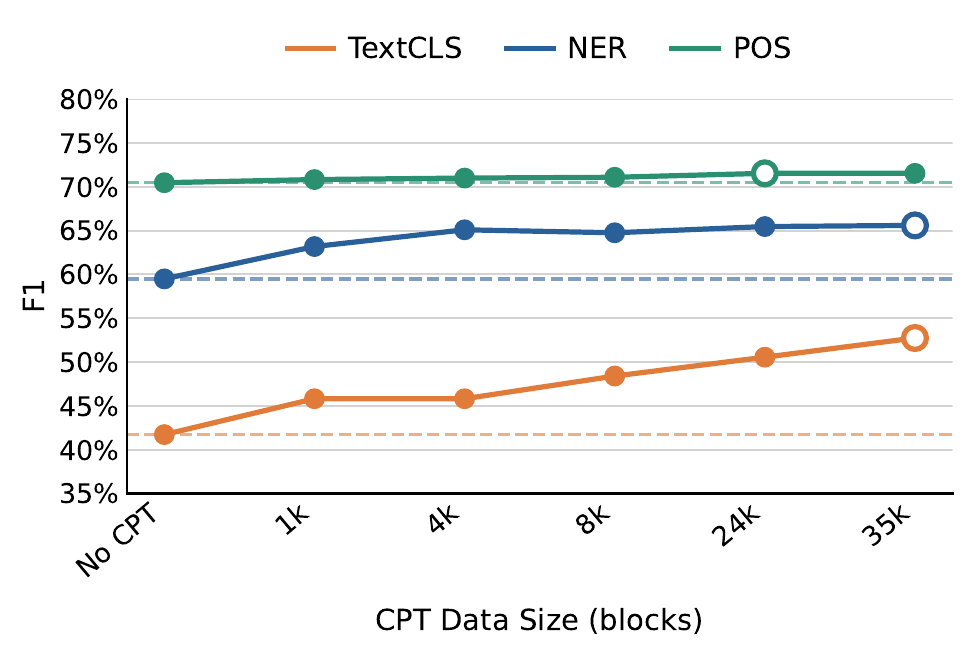}
  \caption{Effect of CPT data scale on downstream task performance.
           Dashed lines indicate No CPT baseline; open circle marks best checkpoint.}
  \label{fig:cpt-ablation}
\end{figure}

Figure \ref{fig:cpt-ablation} shows the performance trends in the three downstream tasks after CPT in Tibetan datasets of varying sizes, starting with \pixel. Overall, the semantic tasks show an upward trends as the scale of CPT training increases, whilst the variation in POS tagging is minimal. The divergence in performance across these tasks is particularly pronounced under ultra-low-resource conditions.

Text classification benefits most from CPT. The macro-average F1 rises from 41.73 (no CPT) to 52.75 (35k \bo), a cumulative improvement of approximately 11 points. Gains are front-loaded: the largest single-step improvement (4.1 points) occurs between no CPT to 1k rendered blocks used for CPT. Performance then plateaus between 1k and 4k (45.82) before resuming steady growth through 8k, 24k and 35k. The curve has not yet reached a clear plateau at 35k, suggesting that given the current data scale, semantic tasks may continue to benefit from additional Tibetan pre-training data.

NER also improves overall, though with a distinct pattern. F1 rises from 59.49 ((no CPT) to 65.59 (35k \bo), a cumulative gain of 6.1 points, again concentrated at early stage. The single-step improvement from no CPT to 1k is 3.7 points. Growth continues through 4k, then shows a small drop at 8k (64.75, down from 65.09) before recovering to 65.46 at 24k and changing negligibly thereafter (65.59 at 35k). This implies NER adaptation may approach saturation around 24k. 

In contrast to Text Classification and NER, POS tagging shows negligible variation across CPT with all data scales, with F1 increasing only 1.1 points, from 70.46 (no CPT) to 71.53 (35k \bo). Tibetan morphological markers, such as case and tense suffixes, are realized as visually distinct characters and thus should be visible by pixel-based models theoretically. However, the higher performance of CANINE-S (77.10) suggests that character-level models may still better exploit such morphological cues for syntactic prediction.

\paragraph{Task Difference.}\label{sec:sem-analysis}
The data scale ablation experiments reveal a consistent pattern of task differentiation: with low-resource CPT, Text Classification, and NER exhibit sensitivity to CPT data scale, whereas POS tagging remains largely unaffected. 

In pixel-based models, one of the core functions of pre-training is to align the models' image representation with the orthographic patterns of the target languages. \citet{tatariya2024pixology} discover that the PIXEL model exhibits a hierarchical division of labor: lower layers primarily capture visual and orthographic features, and deeper layers gradually shift to syntactic and semantic abstraction. Under this view, limited Tibetan CPT may primarily influence the quality of the model's lower level orthographic representations, which could partly explain why semantic tasks benefit more from increasing CPT data. However, task difficulty may play a role and different model types may be better suited for different tasks as discussed above.

Semantic tasks rely heavily on lexical distribution and cross-word semantic associations, and may be more sensitive to increased exposure, explaining their consistent gains with increasing CPT data. In contrast, the syntactic patterns relied upon by POS tagging may benefit less directly from adjustments of Tibetan orthographic representations under the current experimental setup, leading to a relative insensitivity to changes in CPT data scale.

\section{Conclusion}
Using Tibetan as a case study, we explore the feasibility of adapting pixel-based models to low-resource and morphologically complex languages. We propose four rendering-level visual metrics ($PD$, $GC$, $CP$, $CG$) and verify that higher orthographic proximity between auxiliary languages and Tibetan tends to yield better semantic transfer. We demonstrate the feasibility of CPT for adapting pixel models to new scripts, finding that even small-scale CPT data yields noticeable gains on semantic tasks. Through data ablation experiments, we quantify how low-resource we can go while still achieving meaningful adaptation.
Lastly, we find that \pixel benefits more from auxiliary languages pre-training than \pixelm, revealing an inherent trade-off between stronger initial representation capabilities and greater potential for CPT gains.
\vfill

\section*{Limitations}
We only focus on Tibetan and three other low-resource languages written in scripts belonging to the Brahmic family; as well as two  additional languages, Khmer, also written in a Brahmic script, and Chinese, written in Han script. We chose to focus on depth instead of breadth with regards to language choice and analyses. Considering that (1) most languages are unseen by the models, (2) we control for amount of data across languages, and (3) there is a noticeable improvement, even with a very small amount of data, we believe this is valuable nonetheless.

\section*{AI Usage}
We used LLM models for coding assistance, help with grammar, and reviewing a draft.
All the content has been verified and reviewed by the authors, who also take full responsibility. 
All writing is our own.

\section*{Acknowledgments}
This work is based on the first author's master thesis at KU Leuven \cite{zhang2026seeing}.
We thank Kushal Tatariya, Maria Trusca, and the anonymous reviewers for helpful feedback.
W.P. is funded by a KU Leuven Bijzonder Onderzoeksfonds C1 project with reference C14/23/096.
The computational resources and services used were provided by the VSC (Flemish Supercomputer Center), funded by the Research Foundation - Flanders (FWO) and the Flemish Government - department EWI.

\bibliography{bibliography}

\appendix

\section{CPT Details and Architecture}\label{apx:training-details}
We provide the data statistics and hyperparameter configurations for all CPT experiments in this section. Table \ref{tab:cpt_data} provides the data sources and sizes for the four languages used in CPT. Table \ref{tab:cpt_hyperparams} lists the hyperparameter settings shared across all CPT conditions. The overall computational budget is approximately 170-190 GPU hours on NVIDIA H100 GPUs. 

\section{Fine-tuning Details and Architecture}\label{apx:fintuning-details}
We provide fine-tuning dataset statistics and hyperparameter configurations for all downstream task experiments in this section. Table \ref{tab:ft_data} summarizes the dataset resources, splits and evaluation metrics for the three downstream tasks. Table \ref{tab:ft_hyperparams} lists the hyperparameter settings for each task. All models are finetuned from each CPT checkpoint, with the best checkpoint selected based on the validation macro-F1. For POS tagging, we report adjusted macro-F1 that excludes extremely low-frequency tags, \texttt{CONJ}, \texttt{INTJ}, \texttt{SCONJ}, and \texttt{SYM}.

\section{Text Rendering Details}\label{sec:Rendering Details}
All corpora are rendered using \mbox{PangorCairoTextRenderer}, with the Google Noto Sans font family\footnote{\href{https://fonts.google.com/noto}{https://fonts.google.com/noto}} and fallback fonts. The rendered output is a greyscale image with each patch measuring 16*16 pixels. 

\paragraph{CPT Rendering.} A continuous sentence-level rendering strategy is adopted for all four languages: multiple sentences are concatenated and packed into rendering blocks, with each block having a maximum of 511 patches and a minimum of 23 patches. \pixelm \cite{kesen2025multilingual} employed a bi-gram rendering strategy \cite{lotz2023text} to reduce redundant patches and increase patch co-occurrence frequency, thus improving rendering and learning efficiency. However, using such a rendering strategy for Tibetan would not make sense since a single grapheme cluster typically corresponds to between 2 to 8 Unicode code-points, with complex stacked grapheme combinations. This means a bi-gram is not easily defined and may result in splitting of complete orthographic units. Continuous, sentence-level rendering is therefore more appropriate.

\paragraph{Fine-tuning Task Rendering.} For Text Classification, sentences are rendered at the continuous sentence level with a maximum sequence length of 256 patches. For NER and POS tagging, words are rendered as a list to realize word-to-patch alignment, enabling accurate label assignment at the patch level. The maximum sequence length is 196 patches for NER and 256 for POS tagging. These rendering settings generally follow the configurations of \pixel \cite{rust2023language} and \pixelm \cite{kesen2025multilingual}.

\section{Full Experiment Results}\label{apx:full-results}
Table~\ref{tab:full_results} presents the complete experiment results for the downstream tasks under all CPT conditions. The first section covers CPT conditions starting from \pixelm \cite{kesen2025multilingual}, the second section covers Tibetan data-scale ablation experiments starting from \pixel \cite{rust2023language}, and the third section covers auxiliary language experiments starting from \pixel (using 24k rendered blocks for each language). 

\section{41 Language Validation}\label{apx:41 lang}
To validate the four visual similarity metrics beyond our primary case study languages, we sample 41 languages from the Parallel Bible Corpus \cite{mayer-cysouw-2014-creating}, covering six major script groups, Brahmic, Latin, Cyrillic, Han, Hangul and Others, as classified with the ISO 15924 standard. We use the QQ toolkit \cite{poelman2026qq} to verify script metadata for each language, using this to automatically exclude candidates for which the corpus contains only a Latin-script transliteration rather than the language's native script. This automated screening did not flag Malayalam, for which the corpus likewise contains only a transliterated version. We identified this during subsequent manual verification and retain it under the Latin group accordingly (see Table~\ref{tab:visual_similarity_41langs}). Languages are grouped by completeness, full-Bible (n=31) versus New-Testament-only (n=10), and within each group we restrict to a verse-aligned common set (20, 579 verses for the full-Bible group, 7,789 verses for the New-Testament-only group) to control for corpus-content confounds. Preprocessing and rendering follow the same pipeline described in \S\ref{sec:data}.

\begin{figure*}[ht]
    \centering
    \includegraphics[width=\linewidth]{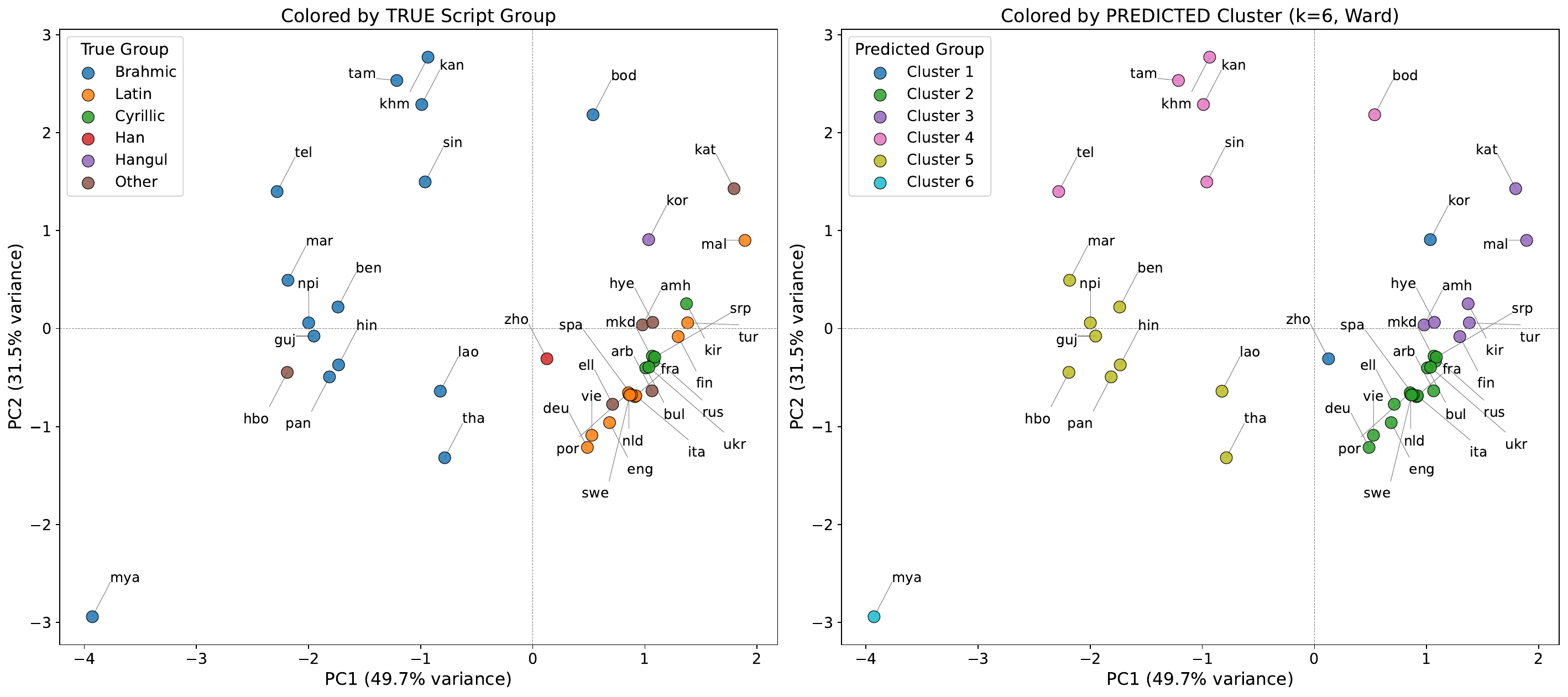}
    \caption{PCA and clustering of the visual metrics. On the left we see the PCA projection and labeling of the ``gold'' groups. On the right we see predicted clusters based on the language vectors (using Ward clustering with $k=6$, we see purity = $0.68$, ARI = $0.33$, NMI = $0.55$, all significantly better than randomization $p < 0.0001$).}
    \label{fig:clustering}
\end{figure*}

For each langauge, we compute the four metrics and represent it as a four-dimensional vector. Figure~\ref{fig:clustering} shows the resulting PCA projection and clustering of these language vectors. Clusters are quite close to what we would expect: Latin and Cyrillic are grouped together, with the Brahmic scripts being quite spread apart.
Part of this is due to the $CP$ metric, which, due to Unicode, is consistently 1 for both Latin and Cyrillic.\footnote{A similar point about inefficiencies in Unicode and UTF-8 was made in \citet{arnett-etal-2024-bit} and \citet{van-der-goot-2026-bytes}.}
Another pattern is the higher overall $GC$ (grapheme clusters per rendered block) being a lot higher for Latin and Cyrillic than the Brahmic scripts.
This is in part due to the increased \emph{width} of characters, leading to fewer characters per block.
For most Brahmic script this is also reflected in higher pixel densities: characters are both wider and more complex.

Based on the resulting language vectors (full values of 41 languages and each language's standardized Euclidean distance to Tibetan reported in Table~\ref{tab:visual_similarity_41langs}), we perform hierarchical clustering using Ward linkage (Euclidean distance), setting k=6 to match the six known script groups, to validate our hypothesis. The resulting clusters align with the true script groups at Purity = 0.68, ARI = 0.33 and NMI = 0.55, all significantly better than a randomized baseline (p $<$ 0.0001); B-cubed evaluation yields Precision = 0.605, Recall = 0.559, F1 = 0.581. This moderate, instead of near-perfect, alignment suggests that the metrics capture finer-grained orthographic similarity beyond script-group labels, rather than simply reproducing the known script classification. 

To examine the discriminative power of each individual metric, we run a univariate analysis of variance for each metric against the six groups. $GC$ ($F=35.12, p < 0.0001$) and $CG$ ($F = 39.83, p < 0.0001$) show significant discriminative power, while $PD$ ($F = 0.57, p = 0.637$) and $CP$ ($F = 0.07, p = 0.977$) do not individually. An ablation study shows that combining all four metrics yields better clustering quality rather than any single metric or partial combination, see Table~\ref{tab:metric-ablation}. This indicates that although $PD$ and $CP$ show relatively weak discriminative power individually, their signal is distributed across interactions with the other metrics, suggesting that combining all four metrics yields more robust clustering than relying on a single metric or a subset.  

\begin{table}[h]
\resizebox{\columnwidth}{!}{
\begin{tabular}{lccc}
\toprule
\textbf{Metric Combination} & \textbf{Purity} & \textbf{ARI} & \textbf{NMI} \\
\midrule
PD only              & 0.585 & 0.182 & 0.351 \\
GC+CP+CG             & 0.659 & 0.267 & 0.472 \\
PD+CG                & 0.683 & 0.319 & 0.506 \\
All four (PD+GC+CP+CG) & \textbf{0.683} & \textbf{0.325} & \textbf{0.548} \\
\bottomrule
\end{tabular}
}
\caption{Ablation of clustering quality ($k=6$, Ward linkage) using different subsets of the four metrics. All four metrics combined yield the best overall clustering quality across all three measures.}
\label{tab:metric-ablation}
\end{table}

\begin{table*}[ht]
\centering
\begin{tabular}{llrr}
\hline
\textbf{Language} & \textbf{Source} & \textbf{Raw Sentences} & \textbf{Rendered Blocks} \\
\hline
Tibetan (\bo) & OPUS \cite{tiedemann2012parallel}  & 714k & 35k \\
& NLLB \cite{costa-jussa2024scaling} \\
Dzongkha (\dz) & FineWeb2 \cite {penedo2025fineweb2a} & 97k & 24k \\
Nepali (\nep) & OPUS \& NLLB & 330k & 24k \\
Burmese (\my) & OPUS \& NLLB & 330k & 24k \\
\hline
\end{tabular}

\caption{CPT data statistics. Raw sentence counts are approximate. 
Rendered block counts reflect the actual number of blocks used 
for CPT after rendering.}
\label{tab:cpt_data}
\end{table*}

\begin{table*}[ht]
\centering
\begin{tabular}{ll}
\hline
\textbf{Parameter} & \textbf{Value} \\
\hline
Image size & (16, 8464, 1) \\
Patch size & 16 \\
Encoder hidden size & 768 \\
Encoder num attention heads & 12 \\
Encoder num layers & 12 \\
Span masking ratio & 0.25 \\
Span masking max length & 6 \\
Optimizer & AdamW \\
Adam $\beta$ & (0.9, 0.999) \\
Adam $\varepsilon$ & 1e-8 \\
Weight decay & 0.05 \\
Base learning rate & 1.5e-4 \\
Effective learning rate & base\_lr $\times$ eff\_bs / 256 \\
Learning rate schedule & Cosine decay \\
Learning rate warmup ratio & 0.05 \\
Training epochs & 15 \\
Per-device batch size & 8 \\
Gradient accumulation steps & 4 \\
Effective batch size & 32 \\
Precision & BF16 \\
Eval samples & 2000$^\dagger$ \\
Seed & 42 \\
\hline
\end{tabular}

\caption{CPT hyperparameter settings for all CPT experiments. 
The architecture follows \citet{rust2023language} and 
\citet{kesen2025multilingual}. 
For data scale experiments with fewer than 24k blocks, 
\texttt{eval\_samples} is scaled proportionally to training set size 
(1k:~100, 4k:~400, 8k:~800).}
\label{tab:cpt_hyperparams}
\end{table*}

\begin{table*}[ht]
\centering
\begin{tabular}{llll}
\hline
\textbf{Parameter} & \textbf{TextCL} & \textbf{NER} & \textbf{POS} \\
\hline
Optimizer & AdamW & AdamW & AdamW \\
Learning rate & 3e-5 & 3e-5 & 3e-5 \\
Weight decay & 0.0 & 0.0 & 0.0 \\
Warmup steps & 100 & 100 & 100 \\
LR scheduler & Linear decay & Linear decay & Linear decay \\
Max epochs & 15 & 15 & 15 \\
Per-device batch size & 8 & 8 & 8 \\
Gradient accumulation steps & 4 & 8 & 8 \\
Effective batch size & 32 & 64 & 64 \\
Max sequence length & 256 & 196 & 256 \\
Dropout probability & — & 0.1 & 0.1 \\
Early stopping patience & 20 & 5 & 5 \\
Metric for best model & Macro F1 & Macro F1 & Macro F1 \\
\hline
\end{tabular}

\caption{Fine-tuning hyperparameters for the three downstream tasks.
All models are finetuned from each CPT checkpoint; the best checkpoint is selected based on validation macro-F1 across all CPT conditions and evaluated on the test set of the respective downstream task.}
\label{tab:ft_hyperparams}
\end{table*}

\begin{table*}[ht]
\centering
\begin{tabular}{llrrrr}
\hline
\textbf{Task} & \textbf{Dataset} & \textbf{Train} & 
\textbf{Val} & \textbf{Test} & \textbf{Metric} \\
\hline
Text Classification & TNCC & 7,420 & 928 & 928 & Macro F1 \\
NER & Tibetan-Mongolian NER & 2,455 & 311 & 306 & Macro F1 \\
POS Tagging & TDTreebank v1.1 & 9,550 & 1,193 & 1,195 & Macro F1 \\
\hline
\end{tabular}

\caption{Downstream fine-tuning dataset statistics. Dataset resources: TNCC \cite{qun2017end}, Tibetan-Mongolian NER \cite{barnett2021named} and TDTreebank v1.1 \cite{quecairang2013treebank}.}
\label{tab:ft_data}
\end{table*}

\begin{table*}[ht]
\centering
\small
\begin{tabular}{lccc}
\hline
\textbf{Condition} & \textbf{NER F1} & \textbf{POS F1} & \textbf{TextCL F1} \\
\hline
\multicolumn{4}{c}{\textit{\pixelm Conditions (mean $\pm$ std, 3 seeds)}} \\
\hline
\pixelm & 69.08{\scriptsize $\pm$0.02} & 70.57{\scriptsize $\pm$0.85} & 44.34{\scriptsize $\pm$0.53} \\
\quad + \bo (35k) & 70.19{\scriptsize $\pm$1.28} & 71.44{\scriptsize $\pm$0.11} & 47.72{\scriptsize $\pm$0.81} \\
\quad + \bo+\dz (35k \bo + 24k \dz) & 69.68{\scriptsize $\pm$0.66} & \textbf{71.76}{\scriptsize $\pm$0.20} & 47.48{\scriptsize $\pm$0.14} \\
\quad + \bo+\my (35k \bo + 24k \my) & 68.26{\scriptsize $\pm$1.80} & 71.15{\scriptsize $\pm$0.54} & 46.34{\scriptsize $\pm$2.40} \\
\quad + \bo+\my (35k \bo + 35k \my) & 68.49{\scriptsize $\pm$0.08} & 71.38{\scriptsize $\pm$0.03} & 46.27{\scriptsize $\pm$0.08} \\
\quad + \bo+\nep (35k \bo + 24k \nep) & 69.38{\scriptsize $\pm$0.57} & 71.26{\scriptsize $\pm$0.48} & 47.75{\scriptsize $\pm$0.13} \\
\quad + allmixed (\bo+\dz+\my+\nep, 107k) & \textbf{70.55}{\scriptsize $\pm$0.02} & 71.72{\scriptsize $\pm$0.12} & \textbf{48.61}{\scriptsize $\pm$0.47} \\
\hline
\multicolumn{4}{c}{\textit{PIXEL --- Data Scale Ablation (mean $\pm$ std, 3 seeds)}} \\
\hline
\pixel & 59.49{\scriptsize $\pm$0.11} & 70.46{\scriptsize $\pm$0.23} & 41.73{\scriptsize $\pm$0.20} \\
\quad + \bo 1k & 63.18{\scriptsize $\pm$0.33} & 70.82{\scriptsize $\pm$0.03} & 45.82{\scriptsize $\pm$0.78} \\
\quad + \bo 4k & 65.09{\scriptsize $\pm$0.33} & 70.99{\scriptsize $\pm$0.04} & 45.82{\scriptsize $\pm$0.19} \\
\quad + \bo 8k & 64.75{\scriptsize $\pm$1.79} & 71.08{\scriptsize $\pm$0.12} & 48.41{\scriptsize $\pm$0.26} \\
\quad + \bo 24k & 65.46{\scriptsize $\pm$0.57} & \textbf{71.53}{\scriptsize $\pm$0.21} & 50.56{\scriptsize $\pm$0.81} \\
\quad + \bo 35k & \textbf{65.59}{\scriptsize $\pm$0.92} & 71.53{\scriptsize $\pm$0.23} & \textbf{52.75}{\scriptsize $\pm$0.10} \\
\hline
\multicolumn{4}{c}{\textit{PIXEL --- Related Language (mean $\pm$ std, 3 seeds)}} \\
\hline
\pixel & 59.49{\scriptsize $\pm$0.11} & 70.46{\scriptsize $\pm$0.23} & 41.73{\scriptsize $\pm$0.20} \\
\quad + \bo 24k & 65.46{\scriptsize $\pm$0.57} & \textbf{71.53}{\scriptsize $\pm$0.21} & 50.56{\scriptsize $\pm$0.81} \\
\quad + \dz 24k & 67.59{\scriptsize $\pm$1.00} & 71.33{\scriptsize $\pm$0.80} & 53.41{\scriptsize $\pm$0.13} \\
\quad + \nep 24k & 64.14{\scriptsize $\pm$1.52} & 70.58{\scriptsize $\pm$0.08} & 47.67{\scriptsize $\pm$0.30} \\
\quad + \my 24k & 63.68{\scriptsize $\pm$0.01} & 70.95{\scriptsize $\pm$0.54} & 46.15{\scriptsize $\pm$0.85} \\
\quad + \bo+\dz (35k \bo + 24k \dz) & \textbf{68.27}{\scriptsize $\pm$0.90} & 71.18{\scriptsize $\pm$0.46} & \textbf{53.71}{\scriptsize $\pm$0.94} \\
\quad + \bo+\my (35k \bo + 24k \my) & 63.27{\scriptsize $\pm$1.41} & 70.75{\scriptsize $\pm$0.56} & 51.12{\scriptsize $\pm$1.26} \\
\quad + \bo+\nep (35k \bo + 24k \nep) & 66.81{\scriptsize $\pm$2.43} & 71.39{\scriptsize $\pm$1.51} & 51.36{\scriptsize $\pm$1.58} \\
\quad + \khm 24k & 60.74{\scriptsize $\pm$1.29} & 70.40{\scriptsize $\pm$1.45} & 51.85{\scriptsize $\pm$0.73} \\
\quad + \zho 24k & 61.63{\scriptsize $\pm$0.45} & 69.77{\scriptsize $\pm$0.86} & 50.85{\scriptsize $\pm$1.42} \\
\hline
\end{tabular}

\caption{Full experiment results across all CPT conditions for three downstream tasks, reported as mean $\pm$ std over three seeds. Bold indicates the best mean performance within each group. The first group reports results with \pixelm{} as the starting point; the second group reports data scale ablation results with \pixel{}; the third group reports auxiliary language CPT results with \pixel{}, each using 24k rendered blocks.}
\label{tab:full_results}
\end{table*}

\begin{table*}
\centering
\small
\begin{tabular}{lllccccc c}
\hline
\textbf{Language} & \textbf{ISO 639-3} & \textbf{ISO 15924} & \textbf{PD} & \textbf{GC} & \textbf{CP} & \textbf{CG} & \textbf{Brahmic} & \textbf{Dist.\ to \texttt{bod}} \\
\hline
Tibetan        & \texttt{bod} & \texttt{Tibt}          & 0.202 & 669.1 & 870.1 & 1.300 & Yes & -- \\
Khmer          & \texttt{khm} & \texttt{Khmr}          & 0.207 & 541.5 & 874.2 & 1.615 & Yes & \cellcolor{blue!15}1.589 \\
Malayalam      & \texttt{mal} & \texttt{Latn}$^\dagger$& 0.186 & 821.5 & 821.5 & 1.000 & No  $^\dagger$ & \cellcolor{blue!18}1.907 \\
Kannada        & \texttt{kan} & \texttt{Knda}          & 0.233 & 524.2 & 799.7 & 1.526 & Yes & \cellcolor{blue!18}2.018 \\
Georgian       & \texttt{kat} & \texttt{Geor}          & 0.236 & 790.4 & 790.4 & 1.000 & No  & \cellcolor{blue!20}2.186 \\
Kyrgyz         & \texttt{kir} & \texttt{Cyrl}          & 0.182 & 768.0 & 768.0 & 1.000 & No  & \cellcolor{blue!20}2.266 \\
Turkish        & \texttt{tur} & \texttt{Latn}          & 0.166 & 775.7 & 775.7 & 1.000 & No  & \cellcolor{blue!21}2.333 \\
Finnish        & \texttt{fin} & \texttt{Latn}          & 0.162 & 768.2 & 768.2 & 1.000 & No  & \cellcolor{blue!22}2.437 \\
Armenian       & \texttt{hye} & \texttt{Armn}          & 0.193 & 731.1 & 731.1 & 1.000 & No  & \cellcolor{blue!23}2.640 \\
Sinhala        & \texttt{sin} & \texttt{Sinh}          & 0.235 & 521.4 & 738.7 & 1.417 & Yes & \cellcolor{blue!23}2.645 \\
Serbian        & \texttt{srp} & \texttt{Cyrl}          & 0.164 & 745.2 & 745.2 & 1.000 & No  & \cellcolor{blue!23}2.648 \\
Russian        & \texttt{rus} & \texttt{Cyrl}          & 0.167 & 741.8 & 741.8 & 1.000 & No  & \cellcolor{blue!23}2.661 \\
Macedonian     & \texttt{mkd} & \texttt{Cyrl}          & 0.162 & 745.3 & 745.3 & 1.000 & No  & \cellcolor{blue!24}2.672 \\
Ukrainian      & \texttt{ukr} & \texttt{Cyrl}          & 0.161 & 741.1 & 741.1 & 1.000 & No  & \cellcolor{blue!24}2.729 \\
Bulgarian      & \texttt{bul} & \texttt{Cyrl}          & 0.163 & 737.3 & 737.3 & 1.000 & No  & \cellcolor{blue!24}2.753 \\
Amharic        & \texttt{amh} & \texttt{Ethi}          & 0.199 & 718.9 & 718.9 & 1.000 & No  & \cellcolor{blue!24}2.794 \\
Tamil          & \texttt{tam} & \texttt{Taml}          & 0.264 & 495.7 & 769.7 & 1.553 & Yes & \cellcolor{blue!25}2.858 \\
Arabic         & \texttt{arb} & \texttt{Arab}          & 0.137 & 753.8 & 756.4 & 1.004 & No  & \cellcolor{blue!25}2.873 \\
Italian        & \texttt{ita} & \texttt{Latn}          & 0.148 & 733.9 & 733.9 & 1.000 & No  & \cellcolor{blue!26}2.954 \\
French         & \texttt{fra} & \texttt{Latn}          & 0.148 & 733.0 & 733.0 & 1.000 & No  & \cellcolor{blue!26}2.964 \\
Spanish        & \texttt{spa} & \texttt{Latn}          & 0.152 & 728.1 & 728.1 & 1.000 & No  & \cellcolor{blue!26}2.972 \\
Portuguese     & \texttt{por} & \texttt{Latn}          & 0.156 & 723.6 & 723.6 & 1.000 & No  & \cellcolor{blue!26}2.982 \\
Swedish        & \texttt{swe} & \texttt{Latn}          & 0.154 & 726.1 & 726.1 & 1.000 & No  & \cellcolor{blue!26}2.983 \\
Dutch          & \texttt{nld} & \texttt{Latn}          & 0.155 & 724.7 & 724.7 & 1.000 & No  & \cellcolor{blue!26}2.986 \\
Telugu         & \texttt{tel} & \texttt{Telu}          & 0.191 & 440.6 & 759.6 & 1.724 & Yes & \cellcolor{blue!27}3.065 \\
Bengali        & \texttt{ben} & \texttt{Beng}          & 0.166 & 480.6 & 720.2 & 1.498 & Yes & \cellcolor{blue!27}3.144 \\
Greek          & \texttt{ell} & \texttt{Grek}          & 0.160 & 707.1 & 707.1 & 1.000 & No  & \cellcolor{blue!27}3.162 \\
Lao            & \texttt{lao} & \texttt{Laoo}          & 0.138 & 563.2 & 715.1 & 1.270 & Yes & \cellcolor{blue!28}3.180 \\
Marathi        & \texttt{mar} & \texttt{Deva}          & 0.149 & 457.0 & 752.5 & 1.646 & Yes & \cellcolor{blue!28}3.215 \\
English        & \texttt{eng} & \texttt{Latn}          & 0.148 & 709.3 & 709.3 & 1.000 & No  & \cellcolor{blue!28}3.260 \\
Nepali         & \texttt{npi} & \texttt{Deva}          & 0.150 & 466.3 & 723.2 & 1.551 & Yes & \cellcolor{blue!29}3.360 \\
Gujarati       & \texttt{guj} & \texttt{Gujr}          & 0.140 & 473.1 & 728.1 & 1.539 & Yes & \cellcolor{blue!29}3.379 \\
Korean         & \texttt{kor} & \texttt{Hang}          & 0.262 & 698.5 & 698.5 & 1.000 & No  & \cellcolor{blue!30}3.454 \\
Hindi          & \texttt{hin} & \texttt{Deva}          & 0.140 & 487.7 & 711.6 & 1.459 & Yes & \cellcolor{blue!30}3.460 \\
Vietnamese     & \texttt{vie} & \texttt{Latn}          & 0.151 & 690.9 & 690.9 & 1.000 & No  & \cellcolor{blue!30}3.463 \\
German         & \texttt{deu} & \texttt{Latn}          & 0.145 & 689.3 & 689.3 & 1.000 & No  & \cellcolor{blue!30}3.550 \\
Punjabi        & \texttt{pan} & \texttt{Guru}          & 0.137 & 481.9 & 705.3 & 1.464 & Yes & \cellcolor{blue!31}3.604 \\
Thai           & \texttt{tha} & \texttt{Thai}          & 0.125 & 569.6 & 683.0 & 1.199 & Yes & \cellcolor{blue!32}3.803 \\
Ancient Hebrew & \texttt{hbo} & \texttt{Hebr}          & 0.095 & 472.1 & 758.2 & 1.606 & No  & \cellcolor{blue!33}3.900 \\
Chinese        & \texttt{zho} & \texttt{Hans}          & 0.247 & 608.3 & 608.3 & 1.000 & No  & \cellcolor{blue!39}4.666 \\
Burmese        & \texttt{mya} & \texttt{Mymr}          & 0.123 & 310.6 & 487.3 & 1.569 & Yes & \cellcolor{blue!60}7.319 \\
\hline
\end{tabular}

\caption{Rendering-level visual similarity metrics (PD, GC, CP, CG; \S\ref{sec:Visual Similarity Metrics}) for 41 languages sampled from the Parallel Bible Corpus \cite[PBC; ][]{mayer-cysouw-2014-creating}, organized by script groups. We use the \textsc{qq} toolkit \cite{poelman2026qq} to gather this metadata. All languages within the same completeness group (full-Bible vs.\ New-Testament-only) are sampled from an identical, verse-aligned set of content, controlling for corpus-content confounds. The rightmost column reports each language's standardized Euclidean distance to Tibetan (\texttt{bod}) over all four metrics. $^\dagger$The corpus contains only a Latin-script transliteration for Malayalam; we verified this directly against the raw corpus file and retain it under Latin accordingly.}
\label{tab:visual_similarity_41langs}
\end{table*}

\end{document}